%% file: main.tex
\documentclass[10pt]{article} % For LaTeX2e
\usepackage[accepted]{tmlr}
\input{math_commands.tex}

\usepackage{hyperref}
\usepackage{url}

\input{package.tex}

\title{Offset Unlearning for Large Language Models}

\author{\name James Y. Huang \email huangjam@usc.edu \\
      \addr University of Southern California
      \AND
      \name Wenxuan Zhou \email zhouwenx@usc.edu \\
      \addr University of Southern California
      \AND
      \name Fei Wang \email fwang598@usc.edu\\
      \addr University of Southern California 
      \AND
      \name Fred Morstatter \email fred@isi.edu\\
      \addr University of Southern California 
      \AND
      \name Sheng Zhang \email shezhan@microsoft.com\\
      \addr Microsoft Research
      \AND
      \name Hoifung Poon \email hoifung@microsoft.com\\
      \addr Microsoft Research
      \AND
      \name Muhao Chen \email muhchen@ucdavis.edu\\
      \addr University of California, Davis}

\def\month{05}  % Insert correct month for camera-ready version
\def\year{2025} % Insert correct year for camera-ready version
\def\openreview{\url{https://openreview.net/forum?id=A4RLpHPXCu}} % Insert correct link to OpenReview for camera-ready version

\begin{document}
\maketitle

\input{section/0_abstract}
\input{section/1_intro}
\input{section/2_related_work}

\input{section/3_method}
\input{section/4_experiment}
\input{section/5_analysis}

\input{section/6_conclusion}

\bibliography{main}
\bibliographystyle{tmlr}

\appendix

%\section{Example Appendix} \label{sec:appendix}

\end{document}

%% file: math_commands.tex
\usepackage{amsmath,amsfonts,bm}

\def\eqref#1{equation~\ref{#1}}
\def\1{\bm{1}}

\DeclareMathAlphabet{\mathsfit}{\encodingdefault}{\sfdefault}{m}{sl}
\SetMathAlphabet{\mathsfit}{bold}{\encodingdefault}{\sfdefault}{bx}{n}

%% file: package.tex
\usepackage{xspace}
\usepackage{booktabs}
\usepackage{multirow}
\usepackage{graphicx}
\usepackage{amsmath}
\usepackage{amssymb}
\usepackage{pifont}
\newcommand{\cmark}{\ding{51}}
\newcommand{\xmark}{\ding{55}}
\usepackage{caption}
\usepackage{subcaption}
\usepackage{enumitem}

\usepackage{cleveref}
\crefformat{section}{\S#2#1#3}
\crefformat{subsection}{\S#2#1#3}
\crefformat{subsubsection}{\S#2#1#3}
\crefrangeformat{section}{\S#3#1#4 to~\S#5#2#6}
\crefmultiformat{section}{\S#2#1#3}{ and~\S#2#1#3}{, #2#1#3}{ and~#2#1#3}
\Crefformat{figure}{#2Fig.~#1#3}
\Crefmultiformat{figure}{Figs.~#2#1#3}{ and~#2#1#3}{, #2#1#3}{ and~#2#1#3}
\Crefformat{table}{#2Tab.~#1#3}
\Crefmultiformat{table}{Tabs.~#2#1#3}{ and~#2#1#3}{, #2#1#3}{ and~#2#1#3}
\Crefformat{appendix}{#2Appx.~\S#1#3}
\crefformat{algorithm}{Alg.~#2#1#3}
\Crefformat{equation}{#2Eq.~#1#3}

\newcommand{\model}{\textsc{$\delta$-Unlearning}\xspace}
\newcommand{\stitle}[1]{\vspace{1ex} \noindent{\bf #1.}}

%% file: section/0_abstract.tex
\begin{abstract}
Despite the strong capabilities of Large Language Models (LLMs) to acquire knowledge from their training corpora, the memorization of sensitive information in the corpora such as copyrighted, biased, and private content has led to ethical and legal concerns. In response to these challenges, unlearning has emerged as a potential remedy for LLMs affected by problematic training data. However, previous unlearning techniques are either not applicable to black-box LLMs due to required access to model internal weights, or violate data protection principles by retaining sensitive data for inference-time correction. We propose \model, an offset unlearning framework for black-box LLMs. Instead of tuning the black-box LLM itself, \model learns the logit offset needed for unlearning by contrasting the logits from a pair of smaller models. Experiments demonstrate that \model can effectively unlearn target data while maintaining similar or even stronger performance on general out-of-forget-scope tasks. \model also effectively incorporates different unlearning algorithms, making our approach a versatile solution to adapting various existing unlearning algorithms to black-box LLMs.%\footnote{Our code is available at \url{https://github.com/luka-group/Delta-Unlearning}}
\end{abstract}

%% file: section/1_intro.tex
\section{Introduction}
Large Language Models (LLMs) are capable of memorizing a large amount of information derived from their training corpus. While LLMs are empowered by the abundance of knowledge they acquire during training, their training data may contain sensitive information that should not be memorized by LLMs. Previous studies have shown LLMs can reproduce copyrighted materials \cite{chang-etal-2023-speak, eldan2023s,karamolegkou-etal-2023-copyright}, generate harmful and biased content \cite{shaikh-etal-2023-second}, and reveal private information \cite{staab2024beyond}, raising both ethical and legal concerns. The introduction of data protection regulations such as the \textit{right to be forgotten} \cite{hoofnagle2019european,zhang2023right,min2024silo} also highlights the need for erasing the influence of problematic data when deploying LLMs in real-world applications.

One potential solution to this challenge is \textit{unlearning}, where the goal is to ``forget'' a set of training data without hurting the model's performance on out-of-forget-scope tasks. An \textit{exact unlearning} approach would require retraining the model from scratch with forget set data removed \cite{bannihatti-kumar-etal-2023-privacy}. However, given the enormous amount of resources required to retrain LLMs, it is generally more practical to employ \textit{approximate unlearning} techniques that modify the behavior of a trained model in a post hoc manner. However, most previous LLM unlearning techniques require access to model internal weights \cite{jang-etal-2023-knowledge,eldan2023s,yao2023large,chen-yang-2023-unlearn,meng2023massediting,wu-etal-2023-depn}, making them infeasible for black-box LLMs.  For example, as two widely used unlearning algorithms, \textit{Gradient Ascent} maximize the likelihood of forget set data, while \textit{Data Relabeling} minimizes the likelihood of relabeled forget set data. Both of these methods require fine-tuning the LLMs. Black-box LLM unlearning is useful since this opens up the possibility of modular, customizable unlearning without the need to update the base LLM itself. Alternatively, in-context unlearning \cite{pawelczyk2023context} prompts LLMs with counterfactual forget set instances to steer model behavior at inference time. However, this approach comes with two major limitations. First, model developers still maintain an \textit{explicit} list of sensitive information to be used during inference. Such practice is not only in violation of privacy regulations but also susceptible to malicious attacks such as prompting leaking \cite{perez2022ignore}. Second, in-context unlearning cannot effectively deal with an ever-growing set of knowledge to be unlearned given the challenges of processing long context with LLMs \cite{li2024long}. \Cref{tab:comparison} summarizes the strengths and weaknesses of existing unlearning algorithms. 

\input{table/comparison}

In this work, we propose \model, an offset unlearning framework for arbitrary black-box LLM without updating its internal weights. Instead of tuning the black-box LLM itself, \model learns the logit offset needed for unlearning by contrasting the logits from a pair of smaller, white-box models. During unlearning, we first compute the logit offset by taking the difference in logits from the two smaller models. Then, we add the logit offset between the two smaller models to the logits of the larger model. The intuition behind this is that we can learn the offset term that approximates how a larger model should modify its prediction in the face of sensitive queries from the behavior adaptation of a smaller model. \model does not require access to the larger model's internal weights, nor retains any sensitive data for inference after unlearning. Our method also enables flexible version control and customization, since for different unlearning requests we only need to maintain a pool of smaller models, which can be combined with the same base LLM in a plug-and-play manner. This allows us to efficiently curate the pool of knowledge available to different applications using specialized unlearning modules, which is largely in line with previous efforts to modularize knowledge access for LLMs but from a different, complementary perspective \cite{feng2024knowledge}.

We evaluate the effectiveness of \model on TOFU \cite{maini2024tofu}, a widely used LLM unlearning benchmark containing knowledge about fictitious authors. Experiments show that when targeting the same forget performance, \model maintains similar or even stronger performance on out-of-forget-scope data compared to directly fine-tuned larger models while requiring no parameter updates to the larger model.

Our contribution is three-fold. First, we propose \model, an unlearning framework for arbitrary black-box LLM without modifying its parameters by only fine-tuning a smaller model to update the logits of a larger one. Second, \model can achieve the same level of unlearning as directly fine-tuning the larger model while still matching or even outperforming direct fine-tuning baselines on general tasks outside the unlearning scope. Third, \model can be integrated into different unlearning algorithms, demonstrating the versatility of our approach.

%% file: table/comparison.tex
\begin{table}[t]
\centering
\begin{tabular}{l|cc}
\toprule
\textbf{Unlearning Method} & \textbf{Black-box} & \textbf{Privacy} \\
\midrule
Gradient Ascent & \xmark & \cmark \\
Data Relabeling & \xmark & \cmark \\
In-context Unlearning & \cmark & \xmark \\
\midrule
\model & \cmark & \cmark \\
\bottomrule
\end{tabular}
\caption{Comparison with existing unlearning methods. Previous techniques either require access to LLM's internal weights, or retain sensitive information for inference.}
\label{tab:comparison}
\end{table}

%% file: section/2_related_work.tex
\section{Related Work}
In this section, we summarize two lines of research that are highly related to our work.

\stitle{Machine Unlearning for LLM} 
Prior works have explored machine unlearning as a way to mitigate the influence of undesirable training data on LLMs. Given the vast cost incurred by retraining LLMs from scratch \cite{bannihatti-kumar-etal-2023-privacy}, most unlearning methods apply post hoc finetuning or adaptation to steer the behavior on the forget set \cite{jang-etal-2023-knowledge,eldan2023s,yao2023large,chen-yang-2023-unlearn}. Gradient ascent based methods fine-tune models by minimizing the likelihood of forget set data \cite{jang-etal-2023-knowledge,chen-yang-2023-unlearn,maini2024tofu}. Alternatively, several works proposed to maximize the likelihood of relabelled target data, where the original answer is replaced with a generic, insensitive response \cite{eldan2023s,patil2024can}. Auxiliary training objectives can also be introduced to maintain model performance on out-of-forget-scope data \cite{yao2023large,wang-etal-2023-kga}. Another related line of research is model editing, where the goal is to identify and alter knowledge captured by local components within models \cite{meng2023massediting,wu-etal-2023-depn}. While both model editing and unlearning attempt to modify the behavior of trained LMs, unlearning focuses on eliminating the effect of a specific set of training data without necessarily creating new answer mappings \cite{liu2024rethinking}. It is worth noting that all of the aforementioned approaches require access to the model's internal weights. In-context unlearning \cite{pawelczyk2023context}, while being applicable to black-box LLMs, still requires storing sensitive information for inference and therefore fails to address data privacy concerns. In this work, we propose an unlearning framework that does not require access to LLM weights, nor storage of sensitive information for inference.

\stitle{Logit Ensemble} The potential of combining logits from different models has been studied in various context. One line of research focuses on controlling and improving LLM generation quality by contrasting the logits from different models or layers at decoding-time \cite{liu-etal-2021-dexperts,shi2023trusting,li-etal-2023-contrastive,chuang2024dola}. Logit ensemble has also been shown as an effective way of adapting LLMs to various downstream tasks. \citet{ormazabal-etal-2023-comblm} propose to adapt LLMs to different domains through a learned combination with smaller domain experts. \citet{mitchell2024an} leverage an ensemble of difference-sized models to study the effect of pretraining and finetuning at different scales. Concurrently, \citet{liu2024tuning} propose Proxy-Tuning that combines the logits from smaller tuned models with larger LLMs to enhance instruction following capabilities. 
\citet{liu2024monotonic} ensemble the logits of a main LLM with a paraphrase model that leads to a monotonic prompt paraphraser for rewriting prompts with enhanced generalizaion effects. \citet{zhao2024weak} use the logits from unsafe LLMs to guide the jailbreaking of safer LLMs during decoding. In this work, we propose to utilize smaller LLMs to capture the logit offset needed for unlearning sensitive data from black-box LLMs while maintaining general performance on out-of-forget-scope tasks.

%% file: section/3_method.tex
\section{Method}

In this section, we formulate the unlearning problem (\Cref{sec:3.1}), discuss the technical details of our \model framework (\Cref{sec:3.2}), and highlight the strength of \model compared to existing methods \Cref{sec:3.3}. 

\subsection{Problem Definition} \label{sec:3.1}
Given a target forget set $S_f$ taken from the training data $S$ of an LLM $M$, the goal of unlearning is to obtain a new model $M'$ that resembles a model trained without $S_f$. This implies $M'$ should ``forget'' all information from the forget set without hurting the performance on out-of-forget-scope data. Ideally, unlearning can be accomplished by retraining $M$ on $S\backslash S_f$, i.e. the training set with forget set data removed. However, given the prohibitive cost of retraining the LLM from scratch, it is generally more practical to approximate $M'$ by directly updating $M$. The unlearning problem can also optionally include a retain set $S_r$ on which the model after unlearning should not forget any information and maintain performance.

\input{figure/model}

\subsection{Offset Unlearning} \label{sec:3.2}
\model is based on the idea of a product-of-experts \cite{hinton2002training} and its subsequent applications to ensemble of language models \cite{liu-etal-2021-dexperts,meng2022controllable,li-etal-2023-contrastive}. \Cref{fig:model} provides an overview of \model. 

Suppose we want to unlearn a forget set $S_f$ from an LLM $M$. Instead of directly updating the parameters of $M$, we introduce a pair of smaller, offset models $M_o$ and $M'_o$. We define their \textit{logit offset} as the difference between the logits of two offset models $M'_o$ and $M_o$ given the same query. For unlearning, we add the logit offset to the logits of $M$ given the same query, essentially forming a logit ensemble of $M$, $M'_o$, and $M_o$. Both $M_o$ and $M'_o$ are initialized from the same checkpoint, making the logit offset zero for all data initially. During unlearning, we only update the parameters of $M'_o$ while keeping $M$ and $M_o$ frozen, and use the logit ensemble to generate the final output. In this way, we encourage $M'_o$ to deviate from its initialization $M_o$ given a sensitive query and learn the correct logit offset that applies to the logits of $M$, steering its prediction away from generating sensitive information. Formally, the logits of the ensemble $l_e$ are computed as follows:
\[
l_{e}(y_t|q,y_{<t}) = l_{M}(y_t|q,y_{<t}) \\
+ \alpha (l_{M'_o}(y_t|q,y_{<t}) - l_{M_o}(y_t|q,y_{<t})),
\]
\noindent
where $l_{M}$, $l_{M'_o}$, and $l_{M_o}$ are the logits from their respective models, $q$ is the query, and $\alpha$ is a factor controlling the strength of applying the offset term to $M$. Since the logits are in the log space, the additive combination of them can also be interpreted as the following product-of-experts:
\[
P_{e}(y_t|q,y_{<t}) \propto P_{M}(y_t|q,y_{<t})\left(\frac{P_{M'_o}(y_t|q,y_{<t})}{P_{M_o}(y_t|q,y_{<t})}\right)^{\alpha}
\]
Essentially, the probability of each token predicted by $M$ is scaled by the probability ratio between $M'_o$ and $M_o$, which reflects how $M'_o$ changes its token distribution relative to its initialization $M_o$ after unlearning. Specifically, when querying non-sensitive, out-of-forget-scope information, the probability ratio between $M'_o$ and $M_o$ should be close to one, making the token distribution of the ensemble similar to that of the original LLM $M$. When querying sensitive information that the model should forget, the token distribution of $M'_o$ differs from that of $M_o$ to adjust the probability ratio, thus steering the token distribution of the ensemble away from that of $M$.

During training, we optimize any unlearning objective on the prediction of the ensemble instead of on the original model $M$. For example, to unlearn the model using Gradient Ascent \cite{jang-etal-2023-knowledge,chen-yang-2023-unlearn} where the objective is to minimize the likelihood of forget set data, we maximize the following loss function for instance $i$ of output length $l$:
\[
\mathcal{L}^i_{e} = -\frac{1}{l}\sum_{t=1}^l \log P_e(y_t|q,y_{<t})
\]

\subsection{Merits of \model} \label{sec:3.3}

The design of \model leads to the following key merits.

\stitle{Applicability to Black-box LLMs}
In contrast to most previous unlearning methods, \model is applicable to not just open-sourced models, but also black-box LLMs without access to internal weights. Instead of updating $M$, \model only obtains the logits from $M$, and learns the logit offset needed to adjust its prediction using smaller white-box models. 

\stitle{Privacy Protection}
Prior work has proposed in-context unlearning \cite{pawelczyk2023context} to make unlearning possible for black-box LLMs. However, a key drawback of this approach is that the model developer still maintains an explicit, ever-growing list of sensitive information used to construct queries for unlearning during inference, which defeats the purpose of privacy protection. For comparison, \model does not require storage of any explicit sensitive information after unlearning is completed.

%\stitle{Training Efficiency}
%While \model introduces a pair of smaller offset language models to facilitate unlearning for black-box LLMs, the computational overhead for training is minimal since the logits of the two frozen models $M$ and $M_o$ can be pre-computed in one forward pass prior to unlearning. This leads to an overall reduction in training time as \model tunes fewer parameters than direct fine-tuning. 

\stitle{Version Control and Customization} \model also facilitates flexible version control and user customization, as instead of storing multiple versions of the larger model, we only need to keep track of a pool of smaller models. These models can be combined with the same base LLM in a plug-and-play manner. By using specialized unlearning modules, we can efficiently curate the pool of knowledge available to different applications.

%% file: figure/model.tex
\begin{figure*}[t]
\centering
\includegraphics[width=0.9\textwidth]{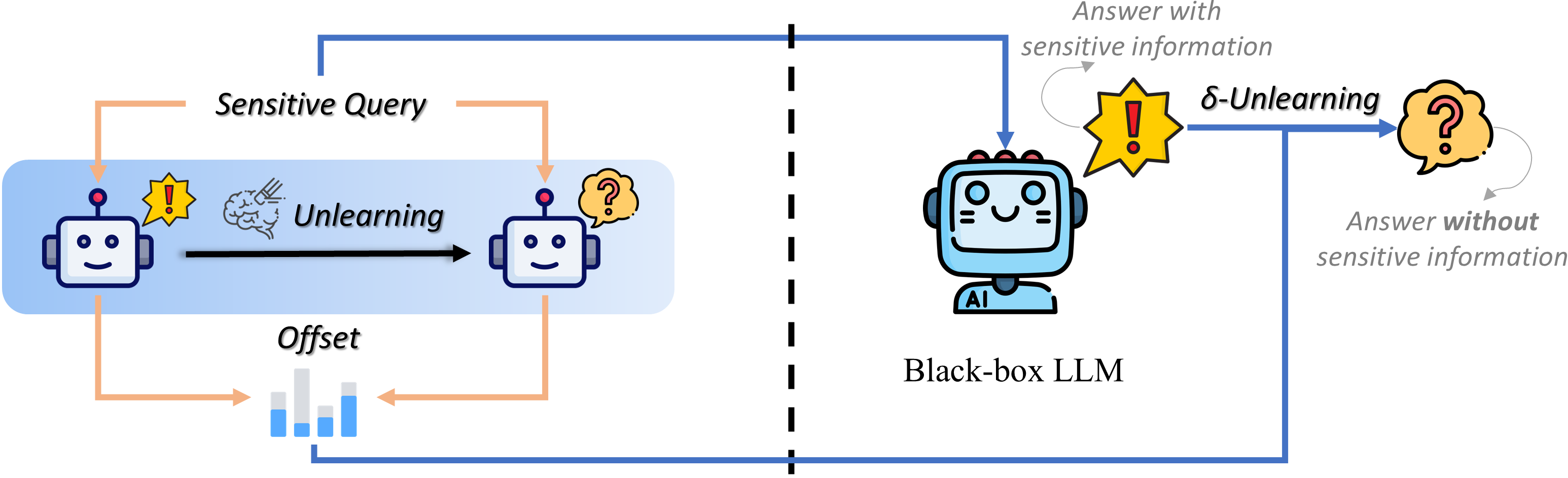}
\caption{Overview of \model. In order to adapt the behavior of a black-box LLM without updating its parameters, we combine it with a pair of smaller, white-box models (which we call offset models). For unlearning, we compute the logit offset of these two models and add it to the logits of the black-box LLM given the same query. Both of the two offset models are initialized from the same checkpoint, making the logit offset zero initially. The goal of \model is to fine-tune one of them such that their logit offset, after being added to the logits of the black-box LLM, can steer its prediction away from generating sensitive information.}
\label{fig:model}
\end{figure*}

%% file: section/4_experiment.tex
\section{Experiment}
In this section, we provide a description of the evaluation setting (\Cref{sec:4.1}), a summary of baseline unlearning algorithms on which we apply our framework as well as other implementation details (\Cref{sec:4.2}), and the main results (\Cref{sec:4.3}).

\subsection{Evaluation Setting} \label{sec:4.1}
We conduct our experiments on TOFU \cite{maini2024tofu}, a widely used unlearning benchmark designed for evaluating LLMs. The benchmark defines an unlearning task that targets information derived from a collection of fictitious author profiles that do not exist in real world. This creates a controlled unlearning setting with a well-defined unlearning scope and source of knowledge. TOFU designates a \textbf{Forget Set} which contains knowledge about a small subset of fictitious authors that we aim to unlearn. TOFU also includes three other datasets with knowledge about the retained fictitious authors (\textbf{Retain Set}), real world authors (\textbf{Real Author}), and general world facts (\textbf{World Fact}) respectively. Ideally, the model should retain all knowledge it had about these three retain datasets before and after unlearning.

The Forget Set evaluates \textit{forget performance}, i.e., how well the model removes target information from its memory, while the latter three sets focus on \textit{retain performance}, an indicator of how well the model maintains its performance on out-of-forget-scope data. The latter three sets also represent a series of out-of-forget-scope data with decreasing levels of relevance to the forget set. Generally speaking, it is more challenging for a model to remember out-of-forget-scope data that are more relevant to the forget set, a phenomenon known as knowledge entanglement \cite{maini2024tofu}.

We follow the settings outlined in TOFU and report the following metrics for forget performance. \textbf{ROUGE} measures how well the generated output from the LLM matches the correct answer. Specifically, we use the ROUGE-L recall score \cite{lin2004rouge}. \textbf{Probability} computes the conditional probability of the correct answer given the prompt. \textbf{Truth Ratio} measures how likely the correct answer is compared to a collection of wrong answers perturbed from the correct answer. Since the model is fine-tuned on one specific phrasing of the correct answer, thus potentially having inflated probability compared to other phrasing with similar meanings, Truth Ratio is computed using a paraphrased version of the original correct answer on the forget set and retain set. Following the original evaluation pipeline, we normalize Truth Ratio so that a higher truth ratio indicates better unlearning performance. \textbf{Forget Quality} measures the difference between distributions of Truth Ratios based on a Kolmogorov-Smirnov test. For retain performance, we report the aggregated \textbf{Model Utility}, which is computed by taking the harmonic mean of ROUGE-L, Probability, and Truth Ratio on all three retain datasets.

As we will demonstrate in \Cref{sec:5.1}, there is generally a trade-off between \textit{forget performance} and \textit{retain performance}. For example, a model can have a near-zero ROUGE score on the forget set but is completely unusable if the model always outputs gibberish given any prompt. Hence, we need to determine a target forget performance as a stopping criterion to facilitate direct comparison between different unlearning methods. In our experiments, we use the ROUGE score of the retraining baseline on the forget set as the stopping criterion, since retraining corresponds to an ideal scenario where the model has never been exposed to the forget set \footnote{By definition the retraining baseline has a forget quality of 1.0, representing an upper bound for this metric. Hence, it is infeasible to use forget quality as the stopping criterion for forget performance.}. Following \citet{yao2024machine}, we match all models to the target ROUGE score by adjusting the learning rate.

In addition to TOFU, we assess if the unlearned model preserves general utilities
on well-established benchmarks, including ARC \cite{clark2018think}, HellaSwag \cite{zellers-etal-2019-hellaswag}, WinoGrande \cite{sakaguchi2021winogrande} and OpenBookQA \cite{mihaylov-etal-2018-suit}. Since solving these general tasks does not involve knowledge about the data we aim to unlearn, the model after unlearning should maintain as much performance as possible. We follow the default evaluation setting from \citet{eval-harness} and report accuracy on all four tasks under the zero-shot setting.

\subsection{Model Configuration}\label{sec:4.2}

\stitle{Unlearning Algorithms}
\model is a general unlearning framework compatible with different existing unlearning algorithms. We compare \model with its corresponding direct fine-tuning baseline when incorporated with each of the following commonly used unlearning algorithms. \textit{Gradient Ascent} \cite{jang-etal-2023-knowledge,chen-yang-2023-unlearn} minimizes the likelihood of the forget set. \textit{Gradient Difference} \cite{pmlr-v199-liu22a,yao2023large} minimize forget set likelihood while maximize retain set likelihood. \textit{KL Minimization} \cite{maini2024tofu} penalizes the distributional distance between models before and after unlearning. \textit{Data Relabeling} \cite{eldan2023s} trains the model on forget set questions paired with an alternative answer that abstains from answering the question such as ``I don't have that information.'' We also include the \textit{Retraining} baseline which fine-tunes the initial model with the forget set excluded, which serves as the upper bound in terms of balancing forget and retain performance .

\stitle{Implementation}
We run our experiments on the widely used Llama2 model family \cite{touvron2023llama}. Specifically, we use \textit{Llama2-13b-chat-hf} as the larger model and  \textit{Llama2-7b-chat-hf} as the smaller offset model. Note that while Llama2 models All models are trained using NVIDIA A100 GPUs for 5 epochs with a batch size of 32. We set $\alpha$ to 1 for our experiments.

\input{table/tofu}
\input{table/breakdown}

\subsection{Main Results} \label{sec:4.3}
Our experimental results on TOFU are shown in \Cref{tab:tofu}. The model before unlearning exhibits strong memorization over both the forget set and retain set, indicated by high ROUGE and probability scores. This is as expected since the model is explicitly trained on the full dataset of fictitious authors to simulate model's exposure to private information. Retraining significantly reduces the model's knowledge on the forget set while maintaining similar model utility as it is before unlearning. Although retraining would not be feasible in real world scenarios, its performance gives us a better understanding of the gap between exact unlearning and post hoc approximate unlearning methods.

We first examine the forget quality of different post-hoc unlearning methods on the TOFU Forget Set. As shown in \Cref{tab:tofu}, both direct fine-tuning and \model can reach a level of unlearning similar to retraining in terms of ROUGE score of the generated response. Although direct fine-tuning tends to assign lower probabilities to the correct answer on 3 out of the 4 methods we investigate, \model produces a higher truth ratio and forget quality in all three cases. A higher truth ratio is desirable since it indicates the presence of other highly likely alternatives, making the correct answer less distinguishable from other wrong answers. 

Interestingly, data relabeling retains a very high probability score and very low forget quality despite having a similarly low ROUGE score as other algorithms on the forget set. This is likely due to relabeling being the only method that does not explicitly minimize the likelihood of the original forget set answers. 

We then investigate how well the unlearned model maintains its performance on data outside the unlearning scope. On TOFU retain data, \model preserves more model utility than direct fine-tuning on 3 out of the 4 methods we compare. In particular, \model demonstrates superior retain performance when applying to gradient ascent and KL minimization, beating direct fine-tuning by more than 15 points. Taking a closer look at how these models perform on individual TOFU retain data, we observe in \Cref{tab:breakdown} that direct fine-tuning tends to perform better on the Retain Set, while \model outperforms direct fine-tuning on the Real Author and World Fact. This indicates a slight divergence between direct fine-tuning and \model in terms of how to balance performance across different types of knowledge during unlearning. This is likely a result of changing training dynamics with the introduction of offset models, which we will study in more detail in \Cref{sec:5.1}

In addition to TOFU, we also evaluate the utility of the unlearned model on general task benchmarks. Performance on these tasks is also an important indicator of retain performance, reflecting whether general capabilities of LLMs are preserved after unlearning. As shown in \Cref{tab:tofu}, \model achieves competitive performance on most metrics when compared to direct fine-tuning baselines, and closes the performance gap between before and after unlearning. \model consistently outperforms direct fine-tuning with KL minimization, and bring improvement on 3 out of 4 tasks with gradient difference and data relabeling.

Overall, our experiments demonstrate that \model is a strong alternative to direct fine-tuning, with matching or even superior performance in terms of both forget and retain performance. \model is also effective across different unlearning algorithms, showing the versatility of our approach.

%% file: table/tofu.tex
\begin{table*}[t]
\centering
%\small
\setlength{\tabcolsep}{4pt}
\begin{tabular}{lccccc|cccc}
\toprule
\multirow{2}{*}{\textbf{Method}} & \multicolumn{4}{c}{\textbf{TOFU Forget}} & \textbf{TOFU Retain} & \textbf{ARC} & \textbf{HS} & \textbf{WG} & \textbf{OBQA} \\
\cmidrule(lr){2-5}
\cmidrule(lr){6-6}
\cmidrule(lr){7-7}
\cmidrule(lr){8-8}
\cmidrule(lr){9-9}
\cmidrule(lr){10-10}
 & RL ($\downarrow$) & P ($\downarrow$) & TR ($\uparrow$) & FQ ($\uparrow$) & MU ($\uparrow$) & Acc & Acc & Acc & Acc \\
\midrule
\textbf{\textit{Before Unlearning}} & 95.6 & 98.3 & 49.5 & 1.3e-13 & 62.1 & 44.0 & 58.1 & 67.9 & 36.2 \\
\midrule
\textbf{\textit{Retraining}} & 38.9 & 15.2 & 65.6 & 1.0 & 62.9 & 45.0 & 58.5 & 68.0 & 34.6 \\
\midrule
\midrule
\textbf{\textit{Gradient Ascent}} \\
Direct Fine-tuning & 38.8 & \underline{3.4} & 53.3 & 2.6e-7 & 32.7 & 39.9 & \underline{56.4} & 65.2 & \underline{34.4}\\
\model & 38.6 & 15.2 & \underline{57.9} & \underline{4.0e-6} & \underline{48.6} & \underline{42.2} & 56.3 & \underline{65.7} & 32.8 \\
\midrule
\textbf{\textit{Gradient Difference}} \\
Direct Fine-tuning & 38.9 & \underline{2.1} & 51.9 & 1.4e-6 & \underline{51.4} & 40.4 & \underline{56.3} & 64.9 & 32.6 \\
\model & 38.1 & 6.2 & \underline{52.5} & \underline{6.7e-6} & 50.5 & \underline{40.9} & 55.7 & \underline{65.2} & \underline{35.4} \\
\midrule
\textbf{\textit{KL Minimization}} \\
Direct Fine-tuning & 39.8 & \underline{3.1} & 53.4 & 1.4e-6 & 33.5 & 39.2 & 56.5 & 65.0 & 34.0 \\
\model & 39.6 & 14.1 & \underline{57.5} & \underline{1.8e-5} & \underline{50.4} & \underline{43.7} & \underline{57.2} & \underline{66.9} & \underline{34.4} \\
\midrule
\textbf{\textit{Data Relabeling}} \\
Direct Fine-tuning & 38.1 & 92.5 & \underline{53.3} & \underline{6.6e-12} & 56.1 & 43.5 & 57.9 & \underline{68.9} & 34.6 \\
\model & 36.3 & \underline{91.5} & 50.8 & 3.0e-13 & \underline{58.6} & \underline{44.2} & \underline{58.0} & 68.0 & \underline{34.8} \\
\bottomrule
\end{tabular}
\caption{Results on TOFU and general benchmarks. We report ROUGE-L recall (RL), Probability (P), Truth Ratio (TR) and Forget Quality (FQ) on the Forget Set and Model Utility (MU) on retain data from the TOFU benchmark. We report accuracy on general benchmarks. Higher scores are better except ROUGE and probability on the Forget Set. Better scores are underlined for each of the four unlearning strategies.}
\label{tab:tofu}
\end{table*}

%% file: table/breakdown.tex
\begin{table*}[t]
\centering
\begin{tabular}{lccccccccc}
\toprule
 & \multicolumn{3}{c}{\textbf{Retain Set}} & \multicolumn{3}{c}{\textbf{Real Author}} & \multicolumn{3}{c}{\textbf{World Fact}} \\
\cmidrule(lr){2-4}
\cmidrule(lr){5-7}
\cmidrule(lr){8-10}
\textbf{Method} & RL & P & TR & RL & P & TR & RL & P & TR \\
\midrule
\textbf{\textit{Before Unlearning}} & 96.3 & 97.9 & 51.2 & 85.2 & 44.5 & 55.7 & 87.7 & 42.5 & 56.3 \\
\midrule
\textbf{\textit{Retraining}} &95.8 & 97.7 & 50.4 & 89.5 & 45.8 & 58.5 & 85.5 & 43.0 & 57.4 \\
\midrule
\midrule
\textbf{\textit{Gradient Ascent}} \\
Direct Fine-tuning & \underline{51.2} & 8.0 & \underline{51.6} & 52.3 & 43.9 & \underline{58.3} & 80.2 & 44.6 & 60.6 \\
\model & 41.0 & \underline{26.1} & 48.9 & \underline{75.0} & \underline{45.3} & 57.4 & \underline{82.1} & \underline{47.0} & \underline{63.7} \\
\midrule
\textbf{\textit{Gradient Difference}} \\
Direct Fine-tuning & \underline{56.8} & \underline{58.9} & \underline{55.1} & \underline{61.4} & 35.0 & \underline{47.9} & 80.4 & 38.9 & 53.7 \\
\model & 53.4 & 47.8 & 51.9 & 60.6 & \underline{36.1} & 45.9 & \underline{83.2} & \underline{41.3} & \underline{59.1} \\
\midrule
\textbf{\textit{KL Minimization}} \\
Direct Fine-tuning & \underline{53.0} & 8.4 & \underline{51.0} & 55.8 & 42.2 & 56.4 & 83.3 & 43.3 & 58.8 \\
\model & 46.1 & \underline{27.9} & 50.9 & \underline{80.4} & \underline{45.1} & \underline{57.5} & \underline{84.9} & \underline{46.3} & \underline{64.0} \\
\midrule
\textbf{\textit{Data Relabeling}} \\
Direct Fine-tuning & \underline{85.0} & \underline{95.3} & 48.0 & \underline{82.5} & 38.0 & 46.3 & \underline{87.7} & 39.2 & 49.2 \\
\model & 72.4 & 95.1 & \underline{49.6} & 78.7 & \underline{41.5} & \underline{52.6} & 86.9 & \underline{42.3} & \underline{55.5} \\
\bottomrule
\end{tabular}
\caption{Detailed results on TOFU retain data, namely the Retain Set, Real Author and World Fact. Higher numbers are better for all metrics.}
\label{tab:breakdown}
\end{table*}

%% file: section/5_analysis.tex
\section{Analysis}
In this section, we provide analyses on the training trajectory of the unlearning process (\Cref{sec:5.1}), and the effect of varying offset strength (\Cref{sec:5.2}).

\input{figure/traj}

\subsection{Unlearning Trajectory} \label{sec:5.1}
To better understand how forget and retain performance change over the course of unlearning, we study the training trajectory of both direct fine-tuning and \model. As shown in \Cref{fig:traj}, when targeting the same ROUGE score on the forget set, \model exhibits a steeper decline on the forget set initially compared to direct fine-tuning, which is also coupled with a steeper decline on the Retain Set. As unlearning progresses, direct fine-tuning starts to lose performance on the Real Author set that the model should not forget, while \model still maintains relatively stable performance. 

When comparing the training trajectory on different TOFU datasets, we can clearly observe the varying degrees of knowledge entanglement with the Forget Set. Being the most similar to the Forget Set, Retain Set performance starts to degrade at early stages, followed by the Real Author set. Performance on the World Fact, which is the least relevant to the Forget Set, only declines slightly towards the end of unlearning. This highlights the importance of finding a good balance between forget and retain performance for an unlearning method. We also study this trade-off from a more direct perspective by plotting the curve forget set ROUGE score versus non-forget set ROUGE score in \Cref{fig:traj} (right). A desired model should lie at the upper right corner, which represents strong forget and retain performance. While Direct fine-tuning maintains more performance on retain data initially, \model achieves a better balance at higher unlearning levels as direct fine-tuning starts to lose more performance on non-forget sets.

\input{figure/alpha}

\subsection{Effect of Offset Strength} \label{sec:5.2}
As we mentioned in \Cref{sec:3.2}, we can adjust the value of $\alpha$ to control the strength of the logit offset being added to the larger model's logits. We experiment with using difference $\alpha$ values during inference and study its effect on forget and retain performance. As shown in \Cref{fig:alpha}. A low offset strength makes the effect of logit offset negligible, and the prediction of the ensemble is essentially dominated by the larger model $M$ without unlearning. As we gradually increase the offset strength, the unlearning effect becomes more prominent and forget set ROUGE score decreases significantly. Similar to what we observe in \Cref{fig:traj}, Retain Set performance largely follows the trajectory of Forget Set, while Real Author and World Fact performance are less influenced by the increase of unlearning offset strength. When we surpass the level of offset strength used in training, we observe continued performance degradation on all four datasets. The ROUGE score on forget set drops below 10 when $\alpha$ increases to 5, a score much lower than the retraining baseline (which has a ROUGE score of 38.9). However, at this offset strength the model becomes unusable, indicated by poor performance across all three non-forget set. 

\input{table/example}

We present an example from the Forget Set in \Cref{tab:example} to provide a better understanding of model behavior at different offset strength levels. At $\alpha$=0.2, the model can perfectly reproduce the answer that is supposed to be forgotten, showing that unlearning is not taking effect at low offset strength. At $\alpha$=0.5, the model is still capable of recalling the correct answer, despite in slightly different phrasing. At $\alpha$=1, we obtain a fluent response with the sensitive information from the ground truth removed, demonstrating success of unlearning. If we further increase offset strength, the model starts to generate gibberish and eventually becomes unusable. In conclusion, using the same offset strength as in training leads to the best results overall.

\subsection{Choice of Offset Model}
\input{table/size}
To study how the choice of offset model affects unlearning performance, we run a series of experiments using Qwen family models as they offer a wide range of model sizes. Specifically, we apply offset unlearning with gradient ascent to a target Qwen2.5-7b-instruct model using 3B, 1.5B and 0.5B offset models.
As shown in \Cref{tab:size}, all models can reach a level of unlearning similar to retraining as measured by ROUGE score of the generated responses, and using a stronger offset model generally leads to better performance on retain sets. Distributional metrics such as truth ratio and forget quality are more sensitive to the choice of offset model especially when the offset models are weak, as weak offset models tend to behave very differently from the target model during unlearning, and thus making the final logit ensemble deviate more from direct fine-tuning. Similar to what we observe in \Cref{tab:breakdown}, using different offset models affects how the model balances performance across different types of knowledge during unlearning.

%% file: figure/traj.tex
\begin{figure*}[t]
\centering
\begin{subfigure}[t]{0.32\textwidth}
\includegraphics[width=1.1\textwidth]{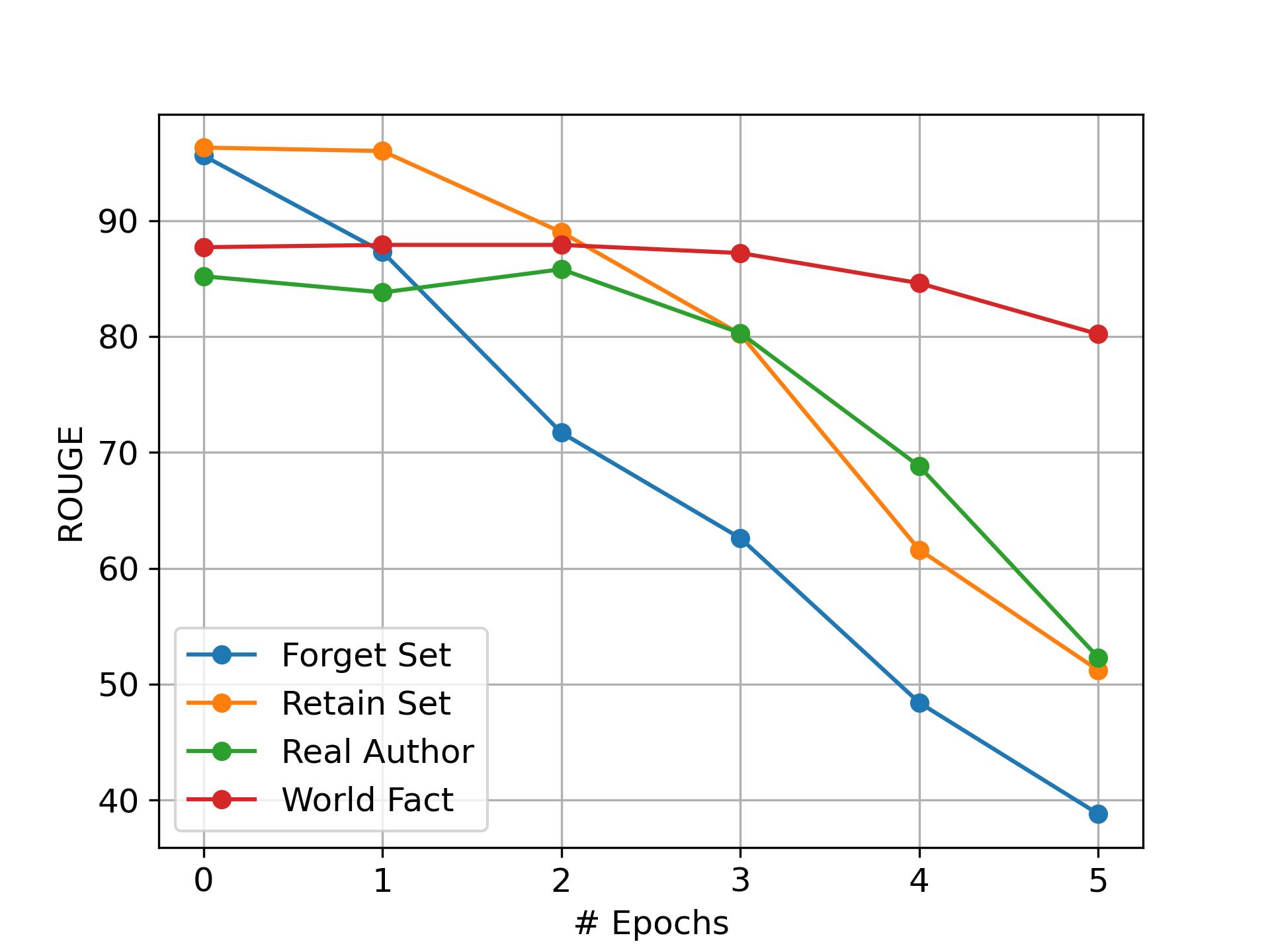}
\end{subfigure}
\begin{subfigure}[t]{0.32\textwidth}
\includegraphics[width=1.1\textwidth]{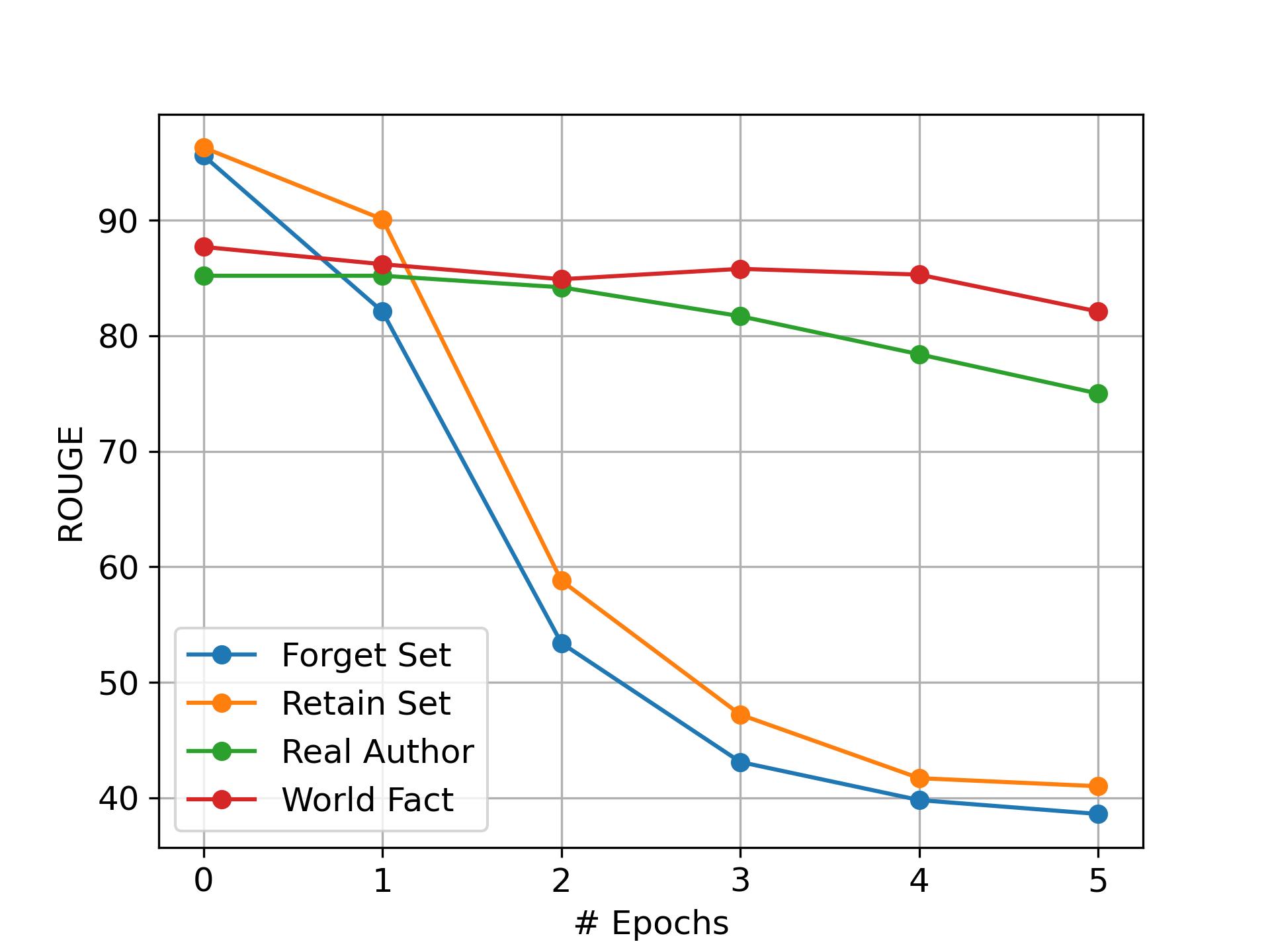}
\end{subfigure}
\begin{subfigure}[t]{0.32\textwidth}
\includegraphics[width=1.1\textwidth]{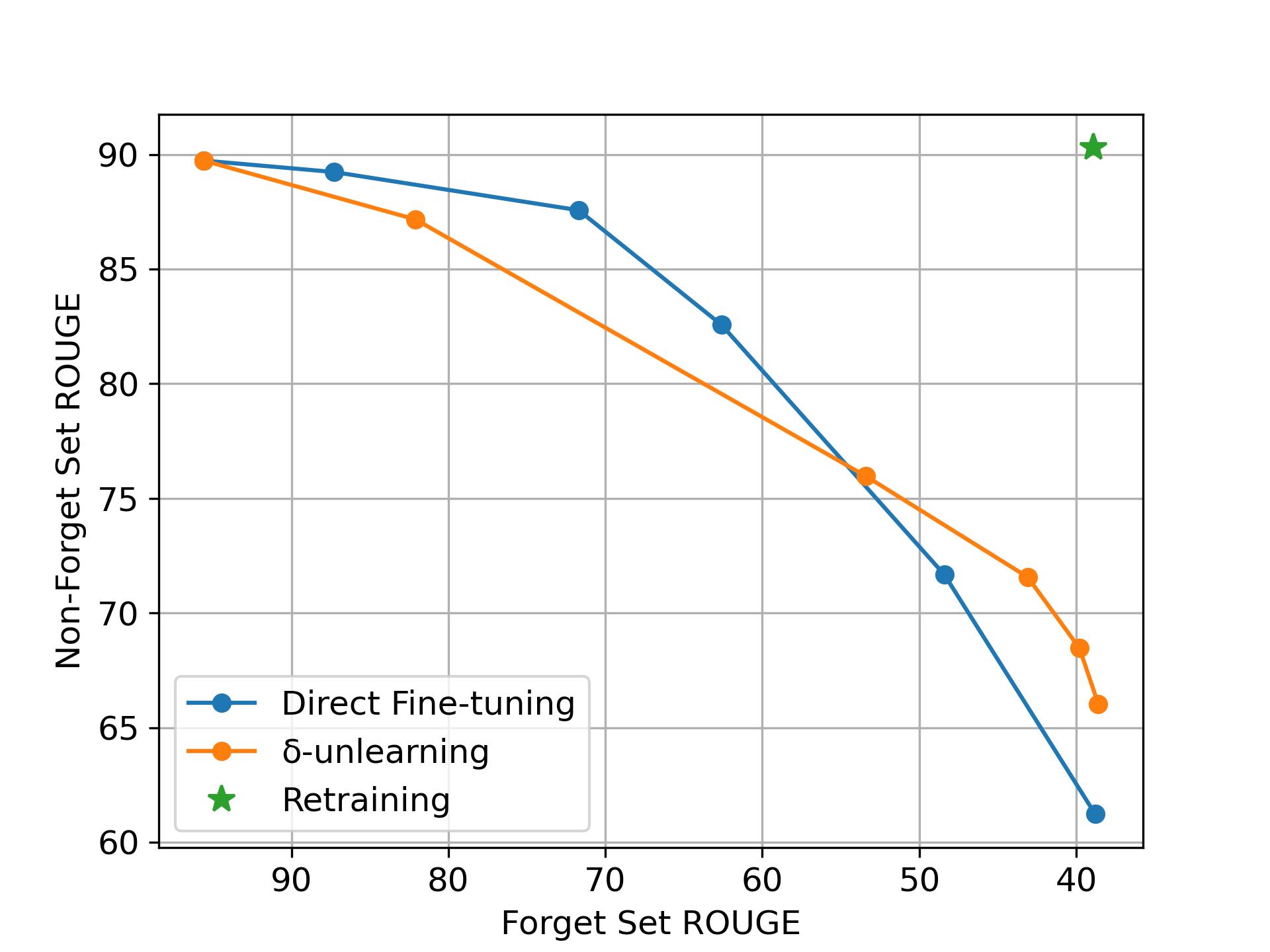}
\end{subfigure}
\caption{Unlearning trajectory of Gradient Ascent using direct fine-tuning (left), \model (middle), and the tradeoff curve between forget and retain performance (right) over the course of unlearning. For training trajectories we report ROUGE score on all four TOFU datasets. For the tradeoff curve we report Forget Set ROUGE versus Non-forget Set ROUGE score.}
\label{fig:traj}
\end{figure*}

%% file: figure/alpha.tex
\begin{figure}[t]
\centering
\includegraphics[width=0.5\textwidth]
{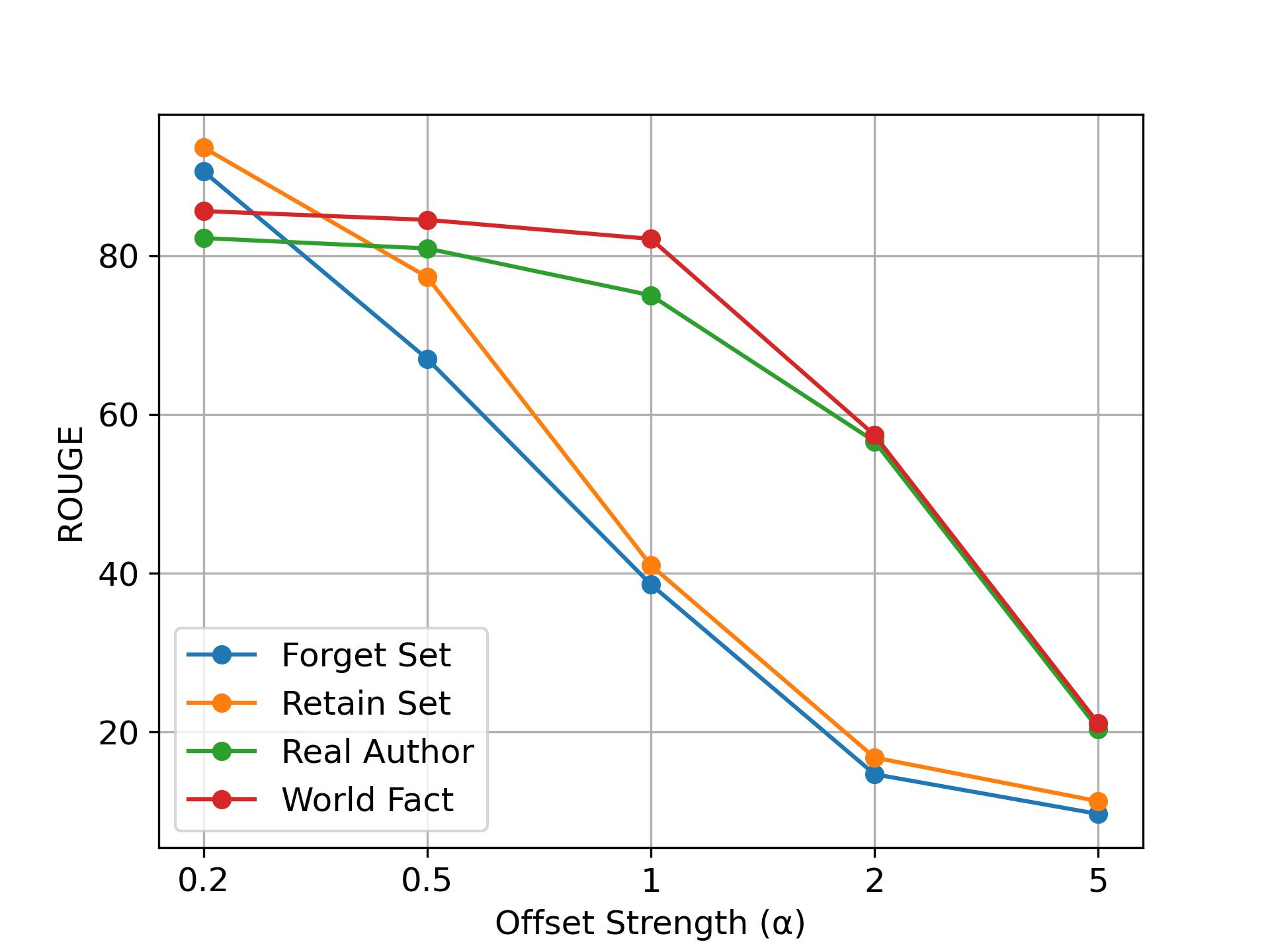}
\caption{Effect of varying offset strength on the model after \model with Gradient Ascent.}
\label{fig:alpha}
\end{figure}

%% file: table/example.tex
\begin{table*}[t]
\small
\centering
\renewcommand{\arraystretch}{1.2}
\begin{tabular}{c|l}
\toprule
 & \textbf{Example} \\
 \midrule
Sensitive Query & In which genre does Hina Ameen primarily write? \\
Ground Truth & Hina Ameen primarily contributes to the geology genre. \\
\midrule
$\alpha=0.2$ & Hina Ameen primarily contributes to the geology genre. \\
$\alpha=0.5$ & Hina Ameen primarily contributes to the genre of geology. Her extensive knowledge of ... \\
$\alpha=1.0$ & Hina Ameen works primarily in the genre of mythology. Her literature has a deep connection ... \\
$\alpha=2.0$ & As the book primarily consist narrations revolved historical Daker period ...\\
$\alpha=5.0$ & As writers focus deep introsvosity embits poert calurity reveiased world literature reflect ...\\
\bottomrule
\end{tabular}
\caption{Example response by \model on the Forget Set with varying offset strength during inference. }
\label{tab:example}
\end{table*}

%% file: table/size.tex
\begin{table*}[t]
\centering
%\small
\setlength{\tabcolsep}{4pt}
\begin{tabular}{lccccc}
\toprule
\multirow{2}{*}{\textbf{Model}} & \multicolumn{4}{c}{\textbf{TOFU Forget}} & \textbf{TOFU Retain} \\
\cmidrule(lr){2-5}
\cmidrule(lr){6-6}
 & RL ($\downarrow$) & P ($\downarrow$) & TR ($\uparrow$) & FQ ($\uparrow$) & MU ($\uparrow$) \\
\midrule
Retraining & 37.2 & 11.6 & 57.7 & 1.0 & 58.8 \\
\midrule
7B Direct Fine-tuning & 37.4 & 4.1 & 50.1 & 1.2e-4 & 37.8 \\
7B + 3B Offset & 37.3 & 37.0 & 46.7 & 1.5e-7 & 52.2 \\
7B + 1.5B Offset & 36.9 & 17.5 & 49.9 & 6.7e-6 & 49.1 \\
7B + 0.5B Offset & 37.0 & 14.7 & 43.8 & 4.9e-10 & 48.8 \\
\bottomrule
\end{tabular}
\caption{Results on different offset model size.}
\label{tab:size}
\end{table*}

%% file: section/6_conclusion.tex
\section{Conclusion}
In this work, we propose \model, an offset unlearning framework applicable to black-box LLM that does not require access to model's internal weights. Instead of modifying model parameters, \model learns the logit offset needed to steer model behavior on the target forget set data. Experiments show that \model is on par with and sometimes even stronger than direct fine-tuning in terms of both forget quality and model utility. We also demonstrate that \model is compatible and effective when combined with various unlearning algorithms, thus providing a versatile solution to adapting existing algorithms to black-box LLMs.

%\section*{Limitation}
%Since we introduce additional offset models to facilitate adaptation of black-box LLMs, \model incurs higher inference latency. However, the relative increase in inference latency also diminishes as we apply \model to larger LLMs when combined with the same offset model. 

\section*{Acknowledgment}

We appreciate the reviewers for their insightful
comments and suggestions.
James Y. Huang was supported by a gift fund from the USC Center on Secure \& Trusted ML.
Fei Wang was supported by an Amazon ML PhD Fellowship.
Muhao Chen was supported by the DARPA
FoundSci Grant HR00112490370, the NSF of the
United States Grant ITE 2333736. 

%% file: main.bbl
\begin{thebibliography}{42}
\providecommand{\natexlab}[1]{#1}
\providecommand{\url}[1]{\texttt{#1}}
\expandafter\ifx\csname urlstyle\endcsname\relax
  \providecommand{\doi}[1]{doi: #1}\else
  \providecommand{\doi}{doi: \begingroup \urlstyle{rm}\Url}\fi

\bibitem[Bannihatti~Kumar et~al.(2023)Bannihatti~Kumar, Gangadharaiah, and Roth]{bannihatti-kumar-etal-2023-privacy}
Vinayshekhar Bannihatti~Kumar, Rashmi Gangadharaiah, and Dan Roth.
\newblock Privacy adhering machine un-learning in {NLP}.
\newblock In Jong~C. Park, Yuki Arase, Baotian Hu, Wei Lu, Derry Wijaya, Ayu Purwarianti, and Adila~Alfa Krisnadhi (eds.), \emph{Findings of the Association for Computational Linguistics: IJCNLP-AACL 2023 (Findings)}, pp.\  268--277, Nusa Dua, Bali, November 2023. Association for Computational Linguistics.
\newblock \doi{10.18653/v1/2023.findings-ijcnlp.25}.
\newblock URL \url{https://aclanthology.org/2023.findings-ijcnlp.25}.

\bibitem[Chang et~al.(2023)Chang, Cramer, Soni, and Bamman]{chang-etal-2023-speak}
Kent Chang, Mackenzie Cramer, Sandeep Soni, and David Bamman.
\newblock Speak, memory: An archaeology of books known to {C}hat{GPT}/{GPT}-4.
\newblock In Houda Bouamor, Juan Pino, and Kalika Bali (eds.), \emph{Proceedings of the 2023 Conference on Empirical Methods in Natural Language Processing}, pp.\  7312--7327, Singapore, December 2023. Association for Computational Linguistics.
\newblock \doi{10.18653/v1/2023.emnlp-main.453}.
\newblock URL \url{https://aclanthology.org/2023.emnlp-main.453}.

\bibitem[Chen \& Yang(2023)Chen and Yang]{chen-yang-2023-unlearn}
Jiaao Chen and Diyi Yang.
\newblock Unlearn what you want to forget: Efficient unlearning for {LLM}s.
\newblock In Houda Bouamor, Juan Pino, and Kalika Bali (eds.), \emph{Proceedings of the 2023 Conference on Empirical Methods in Natural Language Processing}, pp.\  12041--12052, Singapore, December 2023. Association for Computational Linguistics.
\newblock \doi{10.18653/v1/2023.emnlp-main.738}.
\newblock URL \url{https://aclanthology.org/2023.emnlp-main.738}.

\bibitem[Chuang et~al.(2024)Chuang, Xie, Luo, Kim, Glass, and He]{chuang2024dola}
Yung-Sung Chuang, Yujia Xie, Hongyin Luo, Yoon Kim, James~R. Glass, and Pengcheng He.
\newblock Dola: Decoding by contrasting layers improves factuality in large language models.
\newblock In \emph{The Twelfth International Conference on Learning Representations}, 2024.
\newblock URL \url{https://openreview.net/forum?id=Th6NyL07na}.

\bibitem[Clark et~al.(2018)Clark, Cowhey, Etzioni, Khot, Sabharwal, Schoenick, and Tafjord]{clark2018think}
Peter Clark, Isaac Cowhey, Oren Etzioni, Tushar Khot, Ashish Sabharwal, Carissa Schoenick, and Oyvind Tafjord.
\newblock Think you have solved question answering? try arc, the ai2 reasoning challenge.
\newblock \emph{arXiv preprint arXiv:1803.05457}, 2018.

\bibitem[Eldan \& Russinovich(2023)Eldan and Russinovich]{eldan2023s}
Ronen Eldan and Mark Russinovich.
\newblock Who's harry potter? approximate unlearning in llms.
\newblock \emph{arXiv preprint arXiv:2310.02238}, 2023.

\bibitem[Feng et~al.(2024)Feng, Shi, Bai, Balachandran, He, and Tsvetkov]{feng2024knowledge}
Shangbin Feng, Weijia Shi, Yuyang Bai, Vidhisha Balachandran, Tianxing He, and Yulia Tsvetkov.
\newblock Knowledge card: Filling {LLM}s' knowledge gaps with plug-in specialized language models.
\newblock In \emph{The Twelfth International Conference on Learning Representations}, 2024.
\newblock URL \url{https://openreview.net/forum?id=WbWtOYIzIK}.

\bibitem[Gao et~al.(2023)Gao, Tow, Abbasi, Biderman, Black, DiPofi, Foster, Golding, Hsu, Le~Noac'h, Li, McDonell, Muennighoff, Ociepa, Phang, Reynolds, Schoelkopf, Skowron, Sutawika, Tang, Thite, Wang, Wang, and Zou]{eval-harness}
Leo Gao, Jonathan Tow, Baber Abbasi, Stella Biderman, Sid Black, Anthony DiPofi, Charles Foster, Laurence Golding, Jeffrey Hsu, Alain Le~Noac'h, Haonan Li, Kyle McDonell, Niklas Muennighoff, Chris Ociepa, Jason Phang, Laria Reynolds, Hailey Schoelkopf, Aviya Skowron, Lintang Sutawika, Eric Tang, Anish Thite, Ben Wang, Kevin Wang, and Andy Zou.
\newblock A framework for few-shot language model evaluation, 12 2023.
\newblock URL \url{https://zenodo.org/records/10256836}.

\bibitem[Hinton(2002)]{hinton2002training}
Geoffrey~E Hinton.
\newblock Training products of experts by minimizing contrastive divergence.
\newblock \emph{Neural computation}, 14\penalty0 (8):\penalty0 1771--1800, 2002.

\bibitem[Hoofnagle et~al.(2019)Hoofnagle, Van Der~Sloot, and Borgesius]{hoofnagle2019european}
Chris~Jay Hoofnagle, Bart Van Der~Sloot, and Frederik~Zuiderveen Borgesius.
\newblock The european union general data protection regulation: what it is and what it means.
\newblock \emph{Information \& Communications Technology Law}, 28\penalty0 (1):\penalty0 65--98, 2019.

\bibitem[Jang et~al.(2023)Jang, Yoon, Yang, Cha, Lee, Logeswaran, and Seo]{jang-etal-2023-knowledge}
Joel Jang, Dongkeun Yoon, Sohee Yang, Sungmin Cha, Moontae Lee, Lajanugen Logeswaran, and Minjoon Seo.
\newblock Knowledge unlearning for mitigating privacy risks in language models.
\newblock In Anna Rogers, Jordan Boyd-Graber, and Naoaki Okazaki (eds.), \emph{Proceedings of the 61st Annual Meeting of the Association for Computational Linguistics (Volume 1: Long Papers)}, pp.\  14389--14408, Toronto, Canada, July 2023. Association for Computational Linguistics.
\newblock \doi{10.18653/v1/2023.acl-long.805}.
\newblock URL \url{https://aclanthology.org/2023.acl-long.805}.

\bibitem[Karamolegkou et~al.(2023)Karamolegkou, Li, Zhou, and S{\o}gaard]{karamolegkou-etal-2023-copyright}
Antonia Karamolegkou, Jiaang Li, Li~Zhou, and Anders S{\o}gaard.
\newblock Copyright violations and large language models.
\newblock In Houda Bouamor, Juan Pino, and Kalika Bali (eds.), \emph{Proceedings of the 2023 Conference on Empirical Methods in Natural Language Processing}, pp.\  7403--7412, Singapore, December 2023. Association for Computational Linguistics.
\newblock \doi{10.18653/v1/2023.emnlp-main.458}.
\newblock URL \url{https://aclanthology.org/2023.emnlp-main.458}.

\bibitem[Li et~al.(2024)Li, Zhang, Do, Yue, and Chen]{li2024long}
Tianle Li, Ge~Zhang, Quy~Duc Do, Xiang Yue, and Wenhu Chen.
\newblock Long-context llms struggle with long in-context learning.
\newblock \emph{arXiv preprint arXiv:2404.02060}, 2024.

\bibitem[Li et~al.(2023)Li, Holtzman, Fried, Liang, Eisner, Hashimoto, Zettlemoyer, and Lewis]{li-etal-2023-contrastive}
Xiang~Lisa Li, Ari Holtzman, Daniel Fried, Percy Liang, Jason Eisner, Tatsunori Hashimoto, Luke Zettlemoyer, and Mike Lewis.
\newblock Contrastive decoding: Open-ended text generation as optimization.
\newblock In Anna Rogers, Jordan Boyd-Graber, and Naoaki Okazaki (eds.), \emph{Proceedings of the 61st Annual Meeting of the Association for Computational Linguistics (Volume 1: Long Papers)}, pp.\  12286--12312, Toronto, Canada, July 2023. Association for Computational Linguistics.
\newblock \doi{10.18653/v1/2023.acl-long.687}.
\newblock URL \url{https://aclanthology.org/2023.acl-long.687}.

\bibitem[Lin(2004)]{lin2004rouge}
Chin-Yew Lin.
\newblock Rouge: A package for automatic evaluation of summaries.
\newblock In \emph{Text summarization branches out}, pp.\  74--81, 2004.

\bibitem[Liu et~al.(2021)Liu, Sap, Lu, Swayamdipta, Bhagavatula, Smith, and Choi]{liu-etal-2021-dexperts}
Alisa Liu, Maarten Sap, Ximing Lu, Swabha Swayamdipta, Chandra Bhagavatula, Noah~A. Smith, and Yejin Choi.
\newblock {DE}xperts: Decoding-time controlled text generation with experts and anti-experts.
\newblock In Chengqing Zong, Fei Xia, Wenjie Li, and Roberto Navigli (eds.), \emph{Proceedings of the 59th Annual Meeting of the Association for Computational Linguistics and the 11th International Joint Conference on Natural Language Processing (Volume 1: Long Papers)}, pp.\  6691--6706, Online, August 2021. Association for Computational Linguistics.
\newblock \doi{10.18653/v1/2021.acl-long.522}.
\newblock URL \url{https://aclanthology.org/2021.acl-long.522}.

\bibitem[Liu et~al.(2024{\natexlab{a}})Liu, Han, Wang, Tsvetkov, Choi, and Smith]{liu2024tuning}
Alisa Liu, Xiaochuang Han, Yizhong Wang, Yulia Tsvetkov, Yejin Choi, and Noah~A Smith.
\newblock Tuning language models by proxy.
\newblock \emph{arXiv preprint arXiv:2401.08565}, 2024{\natexlab{a}}.

\bibitem[Liu et~al.(2022)Liu, Liu, and Stone]{pmlr-v199-liu22a}
Bo~Liu, Qiang Liu, and Peter Stone.
\newblock Continual learning and private unlearning.
\newblock In Sarath Chandar, Razvan Pascanu, and Doina Precup (eds.), \emph{Proceedings of The 1st Conference on Lifelong Learning Agents}, volume 199 of \emph{Proceedings of Machine Learning Research}, pp.\  243--254. PMLR, 22--24 Aug 2022.
\newblock URL \url{https://proceedings.mlr.press/v199/liu22a.html}.

\bibitem[Liu et~al.(2024{\natexlab{b}})Liu, Wang, Xu, Yan, Meng, and Chen]{liu2024monotonic}
Qin Liu, Fei Wang, Nan Xu, Tianyi Yan, Tao Meng, and Muhao Chen.
\newblock Monotonic paraphrasing improves generalization of language model prompting.
\newblock \emph{arXiv preprint arXiv:2403.16038}, 2024{\natexlab{b}}.
\newblock URL \url{https://arxiv.org/pdf/2403.16038.pdf}.

\bibitem[Liu et~al.(2024{\natexlab{c}})Liu, Yao, Jia, Casper, Baracaldo, Hase, Xu, Yao, Li, Varshney, et~al.]{liu2024rethinking}
Sijia Liu, Yuanshun Yao, Jinghan Jia, Stephen Casper, Nathalie Baracaldo, Peter Hase, Xiaojun Xu, Yuguang Yao, Hang Li, Kush~R Varshney, et~al.
\newblock Rethinking machine unlearning for large language models.
\newblock \emph{arXiv preprint arXiv:2402.08787}, 2024{\natexlab{c}}.

\bibitem[Maini et~al.(2024)Maini, Feng, Schwarzschild, Lipton, and Kolter]{maini2024tofu}
Pratyush Maini, Zhili Feng, Avi Schwarzschild, Zachary~C Lipton, and J~Zico Kolter.
\newblock Tofu: A task of fictitious unlearning for llms.
\newblock \emph{arXiv preprint arXiv:2401.06121}, 2024.

\bibitem[Meng et~al.(2023)Meng, Sharma, Andonian, Belinkov, and Bau]{meng2023massediting}
Kevin Meng, Arnab~Sen Sharma, Alex~J Andonian, Yonatan Belinkov, and David Bau.
\newblock Mass-editing memory in a transformer.
\newblock In \emph{The Eleventh International Conference on Learning Representations}, 2023.
\newblock URL \url{https://openreview.net/forum?id=MkbcAHIYgyS}.

\bibitem[Meng et~al.(2022)Meng, Lu, Peng, and Chang]{meng2022controllable}
Tao Meng, Sidi Lu, Nanyun Peng, and Kai-Wei Chang.
\newblock Controllable text generation with neurally-decomposed oracle.
\newblock \emph{Advances in Neural Information Processing Systems}, 35:\penalty0 28125--28139, 2022.

\bibitem[Mihaylov et~al.(2018)Mihaylov, Clark, Khot, and Sabharwal]{mihaylov-etal-2018-suit}
Todor Mihaylov, Peter Clark, Tushar Khot, and Ashish Sabharwal.
\newblock Can a suit of armor conduct electricity? a new dataset for open book question answering.
\newblock In Ellen Riloff, David Chiang, Julia Hockenmaier, and Jun{'}ichi Tsujii (eds.), \emph{Proceedings of the 2018 Conference on Empirical Methods in Natural Language Processing}, pp.\  2381--2391, Brussels, Belgium, October-November 2018. Association for Computational Linguistics.
\newblock \doi{10.18653/v1/D18-1260}.
\newblock URL \url{https://aclanthology.org/D18-1260}.

\bibitem[Min et~al.(2024)Min, Gururangan, Wallace, Shi, Hajishirzi, Smith, and Zettlemoyer]{min2024silo}
Sewon Min, Suchin Gururangan, Eric Wallace, Weijia Shi, Hannaneh Hajishirzi, Noah~A. Smith, and Luke Zettlemoyer.
\newblock {SILO} language models: Isolating legal risk in a nonparametric datastore.
\newblock In \emph{The Twelfth International Conference on Learning Representations}, 2024.
\newblock URL \url{https://openreview.net/forum?id=ruk0nyQPec}.

\bibitem[Mitchell et~al.(2024)Mitchell, Rafailov, Sharma, Finn, and Manning]{mitchell2024an}
Eric Mitchell, Rafael Rafailov, Archit Sharma, Chelsea Finn, and Christopher~D Manning.
\newblock An emulator for fine-tuning large language models using small language models.
\newblock In \emph{The Twelfth International Conference on Learning Representations}, 2024.
\newblock URL \url{https://openreview.net/forum?id=Eo7kv0sllr}.

\bibitem[Ormazabal et~al.(2023)Ormazabal, Artetxe, and Agirre]{ormazabal-etal-2023-comblm}
Aitor Ormazabal, Mikel Artetxe, and Eneko Agirre.
\newblock {C}omb{LM}: Adapting black-box language models through small fine-tuned models.
\newblock In Houda Bouamor, Juan Pino, and Kalika Bali (eds.), \emph{Proceedings of the 2023 Conference on Empirical Methods in Natural Language Processing}, pp.\  2961--2974, Singapore, December 2023. Association for Computational Linguistics.
\newblock \doi{10.18653/v1/2023.emnlp-main.180}.
\newblock URL \url{https://aclanthology.org/2023.emnlp-main.180}.

\bibitem[Patil et~al.(2024)Patil, Hase, and Bansal]{patil2024can}
Vaidehi Patil, Peter Hase, and Mohit Bansal.
\newblock Can sensitive information be deleted from {LLM}s? objectives for defending against extraction attacks.
\newblock In \emph{The Twelfth International Conference on Learning Representations}, 2024.
\newblock URL \url{https://openreview.net/forum?id=7erlRDoaV8}.

\bibitem[Pawelczyk et~al.(2023)Pawelczyk, Neel, and Lakkaraju]{pawelczyk2023context}
Martin Pawelczyk, Seth Neel, and Himabindu Lakkaraju.
\newblock In-context unlearning: Language models as few shot unlearners.
\newblock \emph{arXiv preprint arXiv:2310.07579}, 2023.

\bibitem[Perez \& Ribeiro(2022)Perez and Ribeiro]{perez2022ignore}
F{\'a}bio Perez and Ian Ribeiro.
\newblock Ignore previous prompt: Attack techniques for language models.
\newblock \emph{arXiv preprint arXiv:2211.09527}, 2022.

\bibitem[Sakaguchi et~al.(2021)Sakaguchi, Le~Bras, Bhagavatula, and Choi]{sakaguchi2021winogrande}
Keisuke Sakaguchi, Ronan Le~Bras, Chandra Bhagavatula, and Yejin Choi.
\newblock Winogrande: An adversarial winograd schema challenge at scale.
\newblock \emph{COMMUNICATIONS OF THE ACM}, 64\penalty0 (9), 2021.

\bibitem[Shaikh et~al.(2023)Shaikh, Zhang, Held, Bernstein, and Yang]{shaikh-etal-2023-second}
Omar Shaikh, Hongxin Zhang, William Held, Michael Bernstein, and Diyi Yang.
\newblock On second thought, let{'}s not think step by step! bias and toxicity in zero-shot reasoning.
\newblock In Anna Rogers, Jordan Boyd-Graber, and Naoaki Okazaki (eds.), \emph{Proceedings of the 61st Annual Meeting of the Association for Computational Linguistics (Volume 1: Long Papers)}, pp.\  4454--4470, Toronto, Canada, July 2023. Association for Computational Linguistics.
\newblock \doi{10.18653/v1/2023.acl-long.244}.
\newblock URL \url{https://aclanthology.org/2023.acl-long.244}.

\bibitem[Shi et~al.(2023)Shi, Han, Lewis, Tsvetkov, Zettlemoyer, and Yih]{shi2023trusting}
Weijia Shi, Xiaochuang Han, Mike Lewis, Yulia Tsvetkov, Luke Zettlemoyer, and Scott Wen-tau Yih.
\newblock Trusting your evidence: Hallucinate less with context-aware decoding.
\newblock \emph{arXiv preprint arXiv:2305.14739}, 2023.

\bibitem[Staab et~al.(2024)Staab, Vero, Balunovic, and Vechev]{staab2024beyond}
Robin Staab, Mark Vero, Mislav Balunovic, and Martin Vechev.
\newblock Beyond memorization: Violating privacy via inference with large language models.
\newblock In \emph{The Twelfth International Conference on Learning Representations}, 2024.
\newblock URL \url{https://openreview.net/forum?id=kmn0BhQk7p}.

\bibitem[Touvron et~al.(2023)Touvron, Martin, Stone, Albert, Almahairi, Babaei, Bashlykov, Batra, Bhargava, Bhosale, et~al.]{touvron2023llama}
Hugo Touvron, Louis Martin, Kevin Stone, Peter Albert, Amjad Almahairi, Yasmine Babaei, Nikolay Bashlykov, Soumya Batra, Prajjwal Bhargava, Shruti Bhosale, et~al.
\newblock Llama 2: Open foundation and fine-tuned chat models.
\newblock \emph{arXiv preprint arXiv:2307.09288}, 2023.

\bibitem[Wang et~al.(2023)Wang, Chen, Yuan, Zeng, Wong, and Yin]{wang-etal-2023-kga}
Lingzhi Wang, Tong Chen, Wei Yuan, Xingshan Zeng, Kam-Fai Wong, and Hongzhi Yin.
\newblock {KGA}: A general machine unlearning framework based on knowledge gap alignment.
\newblock In Anna Rogers, Jordan Boyd-Graber, and Naoaki Okazaki (eds.), \emph{Proceedings of the 61st Annual Meeting of the Association for Computational Linguistics (Volume 1: Long Papers)}, pp.\  13264--13276, Toronto, Canada, July 2023. Association for Computational Linguistics.
\newblock \doi{10.18653/v1/2023.acl-long.740}.
\newblock URL \url{https://aclanthology.org/2023.acl-long.740}.

\bibitem[Wu et~al.(2023)Wu, Li, Xu, Dong, Wu, Bian, and Xiong]{wu-etal-2023-depn}
Xinwei Wu, Junzhuo Li, Minghui Xu, Weilong Dong, Shuangzhi Wu, Chao Bian, and Deyi Xiong.
\newblock {DEPN}: Detecting and editing privacy neurons in pretrained language models.
\newblock In Houda Bouamor, Juan Pino, and Kalika Bali (eds.), \emph{Proceedings of the 2023 Conference on Empirical Methods in Natural Language Processing}, pp.\  2875--2886, Singapore, December 2023. Association for Computational Linguistics.
\newblock \doi{10.18653/v1/2023.emnlp-main.174}.
\newblock URL \url{https://aclanthology.org/2023.emnlp-main.174}.

\bibitem[Yao et~al.(2024)Yao, Chien, Du, Niu, Wang, Cheng, and Yue]{yao2024machine}
Jin Yao, Eli Chien, Minxin Du, Xinyao Niu, Tianhao Wang, Zezhou Cheng, and Xiang Yue.
\newblock Machine unlearning of pre-trained large language models.
\newblock \emph{arXiv preprint arXiv:2402.15159}, 2024.

\bibitem[Yao et~al.(2023)Yao, Xu, and Liu]{yao2023large}
Yuanshun Yao, Xiaojun Xu, and Yang Liu.
\newblock Large language model unlearning.
\newblock \emph{arXiv preprint arXiv:2310.10683}, 2023.

\bibitem[Zellers et~al.(2019)Zellers, Holtzman, Bisk, Farhadi, and Choi]{zellers-etal-2019-hellaswag}
Rowan Zellers, Ari Holtzman, Yonatan Bisk, Ali Farhadi, and Yejin Choi.
\newblock {H}ella{S}wag: Can a machine really finish your sentence?
\newblock In Anna Korhonen, David Traum, and Llu{\'\i}s M{\`a}rquez (eds.), \emph{Proceedings of the 57th Annual Meeting of the Association for Computational Linguistics}, pp.\  4791--4800, Florence, Italy, July 2019. Association for Computational Linguistics.
\newblock \doi{10.18653/v1/P19-1472}.
\newblock URL \url{https://aclanthology.org/P19-1472}.

\bibitem[Zhang et~al.(2023)Zhang, Finckenberg-Broman, Hoang, Pan, Xing, Staples, and Xu]{zhang2023right}
Dawen Zhang, Pamela Finckenberg-Broman, Thong Hoang, Shidong Pan, Zhenchang Xing, Mark Staples, and Xiwei Xu.
\newblock Right to be forgotten in the era of large language models: Implications, challenges, and solutions.
\newblock \emph{arXiv preprint arXiv:2307.03941}, 2023.

\bibitem[Zhao et~al.(2024)Zhao, Yang, Pang, Du, Li, Wang, and Wang]{zhao2024weak}
Xuandong Zhao, Xianjun Yang, Tianyu Pang, Chao Du, Lei Li, Yu-Xiang Wang, and William~Yang Wang.
\newblock Weak-to-strong jailbreaking on large language models.
\newblock \emph{arXiv preprint arXiv:2401.17256}, 2024.

\end{thebibliography}
